\documentclass[letterpaper, 10 pt, conference]{ieeeconf}  

\IEEEoverridecommandlockouts                              

\usepackage{graphicx} 
\usepackage{amsmath} 
\usepackage{paralist}
\usepackage{xcolor}
\usepackage{algorithm}
\usepackage{algpseudocode}
\usepackage{subcaption}
\usepackage{xspace}
\usepackage{units}
\usepackage{etoolbox} 
\usepackage{hyperref}

\title{\LARGE \bf
\smac: Symbiotic Multi-Agent Construction}

\author{
  Caleb Wagner$^{*,1}$,
  Neel Dhanaraj$^{*,1}$,
  Trevor Rizzo$^{1}$,\\
  Josue Contreras$^{1}$,
  Hannan Liang$^{1}$,
  Gregory Lewin$^{1}$,
  Carlo Pinciroli$^{1}$
\thanks{
    $^{*}$C.\ Wagner and N.\ Dhanaraj contributed equally to this work.}
\thanks{
    $^{1}$C.\ Wagner, N.\ Dhanaraj, T.\ Rizzo, J.\ Contreras, B.\ Liang, G.\ Lewin and C.\ Pinciroli are with
    the Department of Robotics Engineering, Worcester Polytechnic Institute, MA 01609, USA.
    {\tt \{cwagner, ndhanaraj, tarizzo, hliang2, jdcontrerasalbuj, glewin, cpinciroli\}@wpi.edu}.}}

\newcommand{\smac}{SMAC\xspace}

\newbool{extendedtext}
\setbool{extendedtext}{false}

\begin{document}

\maketitle
\thispagestyle{empty}
\pagestyle{empty}

\begin{abstract}
We present a novel concept of a heterogeneous, distributed platform for autonomous 3D construction. The platform is composed of two types of robots acting in a coordinated and complementary fashion:
\begin{inparaenum}[(i)]
\item A collection of communicating \emph{smart construction blocks} behaving as a form of growable smart matter, and capable of planning and monitoring their own state and the construction progress; and
\item A team of \emph{inchworm-shaped builder robots} designed to navigate and modify the 3D structure, following the guidance of the smart blocks.
\end{inparaenum}
We describe the design of the hardware and introduce algorithms for navigation and construction that support a wide class of 3D structures. We demonstrate the capabilities of our concept and characterize its performance through simulations and real-robot experiments.
\end{abstract}

\section{INTRODUCTION}



Multi-robot systems promise solutions for construction at much larger scales than the robots themselves. With many robots working in parallel, construction can be completed in a fast and efficient manner. Decentralization might also result in a robust solution with no single point of failure, paving the way for uncrewed construction of colonies in adverse environments such as the Moon or Mars.

The design space of possible approaches to collective construction is vast, and it encompasses a careful co-design of both hardware, software, and fabrication \cite{CollectiveConstruction}. A particularly important open problem is how to coordinate the construction process autonomously. In this paper, we frame this problem as the study of \textit{how intelligence should be distributed across the system}. 

To answer this research question, we explore a novel concept for multi-robot construction that comprises two types of robots. The first is a collection of \emph{smart construction blocks} behaving as a form of growing smart matter. By communicating, the smart blocks monitor the state of the structure and its construction progress in a decentralized fashion. The second type of robot is the \emph{builder}, unaware of the global target structure but able to manipulate and transport blocks in 3D thanks to its inchworm-shaped body. Builders perform localization and path planning by communicating with the smart blocks, which perform the necessary computation. The two types of robots constitute a \emph{symbiotic} system in that their continuous collaboration is necessary for success. For this reason, we called our approach \emph{SMAC} --- \emph{Symbiotic Multi-Agent Construction}.

We envision our system as a technology for in-orbit or planetary construction. In addition, while the smart blocks could be used as actual construction material, they would also be suitable as \emph{smart scaffolding} \cite{komendera2011assembly}. The latter idea would enable reusable, cost-effective technology for remote, large-scale, autonomous construction.

\begin{figure}
    \centering
    \includegraphics[width=.8\columnwidth]{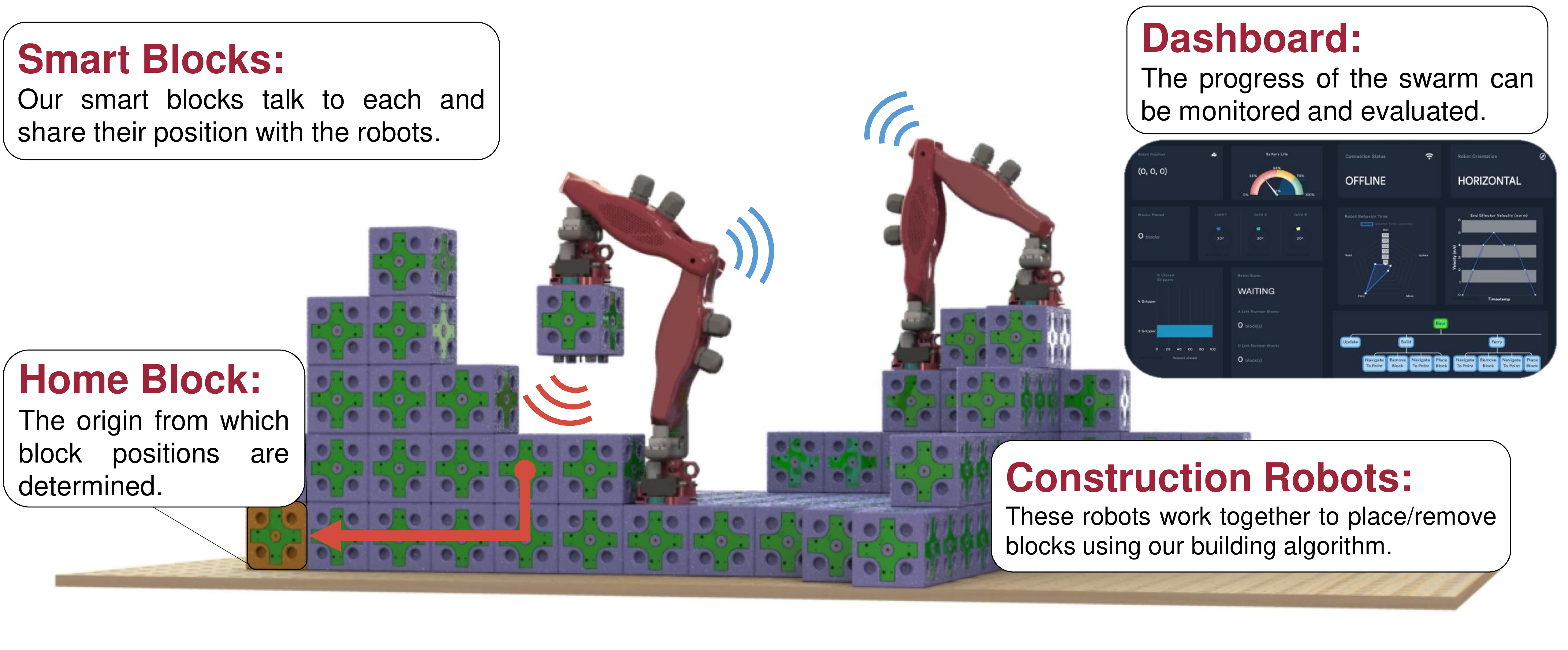}
    \caption{Overview of proposed concept.}
    \label{fig:system_architecture}
\end{figure}

\textbf{Related work.} The idea of performing construction in which a robot interacts with the environment is inspired by stigmergy in social insects \cite{theraulaz1999brief}. Several approaches have been proposed in this context, both in simulation \cite{theraulaz1995coordination,grushin2006stigmergic,werfel2007collective,Pinciroli:ANTS2018} and with real robotic platforms \cite{martinoli1999understanding,napp2014distributed,soleymani2015bio,SROCS}. In particular, SROCS introduces the concept of \textit{smart blocks} capable of communicating with a builder robot, and of storing and retrieving information. \smac differs from SROCS in two crucial ways:
\begin{inparaenum}[\it (i)]
\item in \smac, smart blocks communicate with each other, acting as an intelligent swarm;
\item in SROCS, the builder robot is a crane incapable of traversing the ongoing structure, making it impossible to build structures beyond the builder's own size.
\end{inparaenum}
The problem of navigating and localizing within an ongoing structure has been the subject of several influential works. TERMES \cite{petersen_termes} pioneered a concept in which a wheg-equipped robot, co-designed with a convex block, constructs 2.5D structures by block deposition and removal. The robot follows a predefined plan calculated by an optimization algorithm \cite{deng2019scalable}, with limited capabilities to recover from placing errors. Bill-E \cite{Bill-E} is a bipedal, inchworm-shaped robot designed to navigate and build versatile 3D structures composed of special blocks. Both TERMES and Bill-E handle passive blocks, and localization is achieved by tracking the number of blocks covered since the entry point of the structure.

\textbf{Novelty and contributions.}
To the best of our knowledge, this work is the first to explore the co-design space of collective construction with symbiotic robot teams comprising growing active matter and mobile robots. In this domain, our paper provides four main contributions:
\begin{enumerate}
\item We propose the co-design of two types of robots, both of which present novel challenges and solutions;
\item We introduce planning algorithms for both types of robots that achieve 3D construction and navigation;
\item We characterize the class of structures that are achievable through our system;
\item We demonstrate the effectiveness of our approach through simulations (particularly regarding scalability) and real-world experiments (to validate the hardware design).
\end{enumerate}

The paper is organized as follows. In Sec.\ \ref{sec:problem_statement} we list the main requirements. In Sec.\ \ref{sec:robot_platforms} we illustrate the hardware design, and in Sec.\ \ref{sec:planning_navigation} we describe our algorithms for planning and navigation. We report the results of the evaluation of our approach in Sec.\ \ref{sec:experimental_evaluation} and conclude the paper in Sec.\ \ref{sec:conclusions}.

\section{PROBLEM STATEMENT}
\label{sec:problem_statement}

\smac is a multi-robot construction platform designed with the intent of exploring novel questions regarding large-scale, multi-robot, 3D construction. In particular, we aim to separate the planning/monitoring logic involved with multi-robot construction from the physical aspects of construction, such as navigation and block placement. This will enable the study of novel research questions centered around the concept of \textit{``active'' stigmergy}, in which simple, reactive robots build smart structures composed of modules capable of dynamic monitoring and (re)planning.

For the purposes of this work, we consider three high-level design requirements to achieve this vision:
\begin{enumerate}
\item We want our builders to navigate over challenging 3D structures. These robots must also be capable of manipulation/placement of building blocks in any part of the structure.
\item We want our smart blocks to locally compute and communicate information pertinent to the ongoing construction task. Communication must occur both among blocks and between a block and an attached inchworm.
\item Finally, we aim to demonstrate the potential of this kind of distributed, spatial intelligence by devising algorithms that achieve construction in an efficient and scalable manner.
\end{enumerate}

\section{ROBOT PLATFORMS}
\label{sec:robot_platforms} 

\subsection{Inchworm Builder and Smart Block Co-Design}
\label{sec:co_design} 
We start our presentation by defining the parameters and decisions that dictate the inchworm and block co-design. 

\textbf{Smart block side length.}
The first design parameter is the side length $L$ of the smart blocks, which we consider the ``unit'' of the lattice constituting any target structure. A large side length would allow for better electronics and mechanical capabilities for a block, but it would also increase the size and weight of the inchworm. A short length, on the other hand, would allow for a smaller, lighter inchworm, while limiting the capabilities of the smart block.

\textbf{Inchworm-block docking interface.}
The docking interface between an inchworm and a block face dictates how the inchworm climbs a structure and manipulates a block, as well as how the blocks assemble to become a structure. We selected a 4-dowel pin-to-hole design for two reasons:
\begin{inparaenum}[\it (i)]
\item This type of interface can sustain sufficient shear loads; and
\item it fosters alignment, which is beneficial for both inchworms and newly placed blocks onto the structure. For the robots and blocks to exert an attachment force onto the structure, we used an actuated threaded screw. This interface is shown in Fig. \ref{fig:interface_cad}.
\end{inparaenum}

\textbf{Inchworm-block communication.}
The communication interface between an inchworm and a block occurs between the inchworm feet and any face of a block. This allows the inchworm to both communicate and localize itself on the structure without ambiguity.

\subsection{Inchworm Builders}

In the literature, robots such as whegged \cite{petersen_termes}, aerial \cite{stuart_swarm_const_drone}, quadrupedal \cite{jpl_lemur}, and bipedal robots \cite{Bill-E} have demonstrated the traversal of simple 3D structures to place building material with varying levels of dexterity and control complexity \cite{CollectiveConstruction}. In the quest for a versatile design with low control complexity, we identified an inchworm-shaped platform as a satisfactory solution \cite{Bill-E}. An inchworm can traverse over almost all block structures. Furthermore, an inchworm robot can pick and place blocks onto structures at any position and orientation in their workspace. This capability does not compromise other important factors such as weight and cost.

\begin{figure}[t]
\centering
\includegraphics[width=.9\columnwidth]{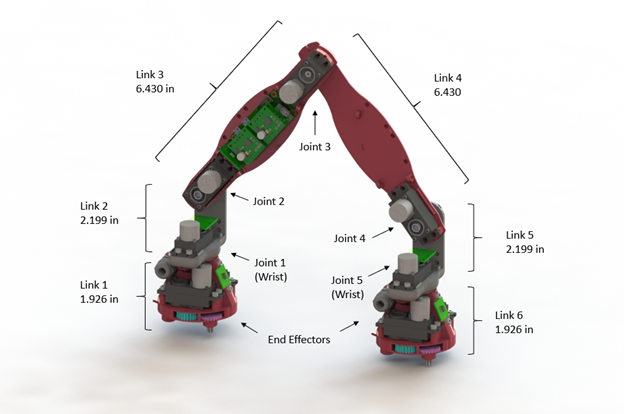}
\caption{CAD design of the inchworm builder.}
\label{fig:inchworm_cad}
\end{figure}

\textbf{Inchworm design summary.} Our inchworm robot is formally defined as a symmetric, bipedal 6-link serial-linkage robot with 5 rotational joints (see Fig.\ \ref{fig:inchworm_cad}). Two of these joints are wrist joints located at each end effector, which enables the inchworm to manipulate the blocks' positions and orientation along 2 axes of rotation and traverse 3D structures.

\begin{figure}[t]
\centering
\includegraphics[width=.9\columnwidth]{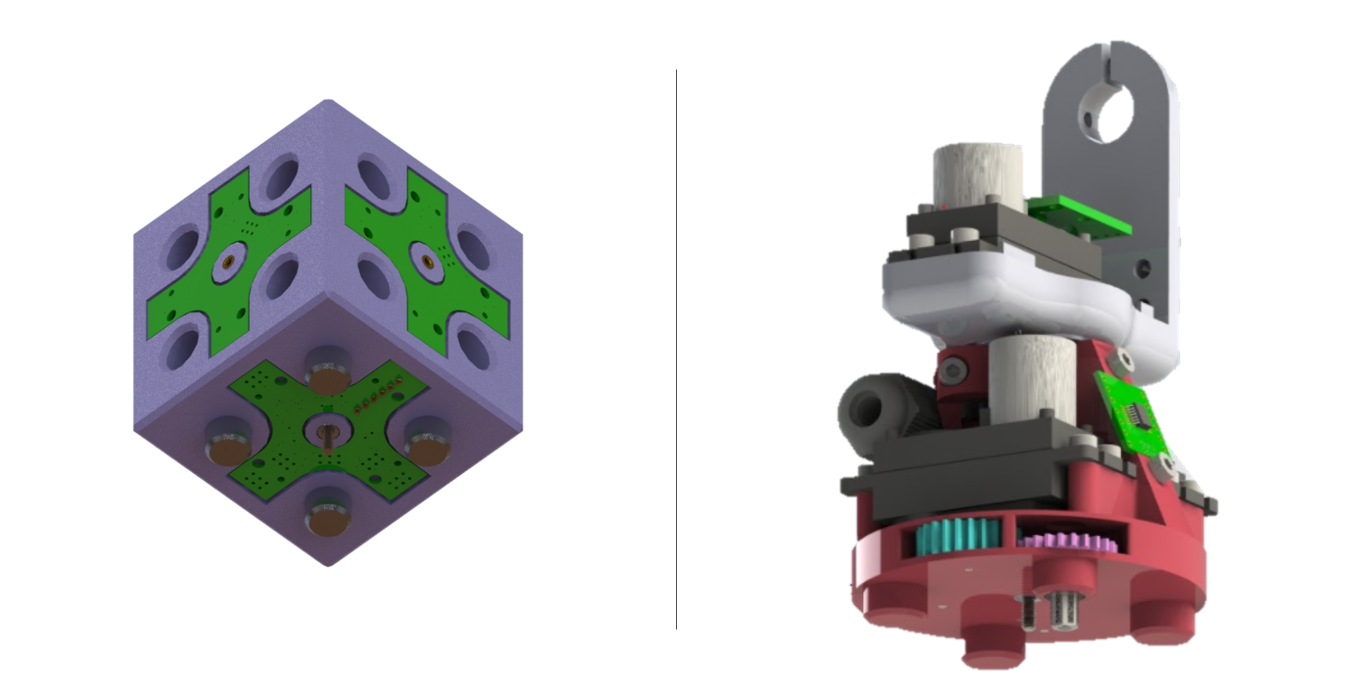}
\caption{CAD model of the inchworm end effector and smart block. This figures shows the common interface mechanism used by both designs.}
\label{fig:interface_cad}
\end{figure}

\begin{figure}[t]
    \centering
    \includegraphics[width=.9\columnwidth]{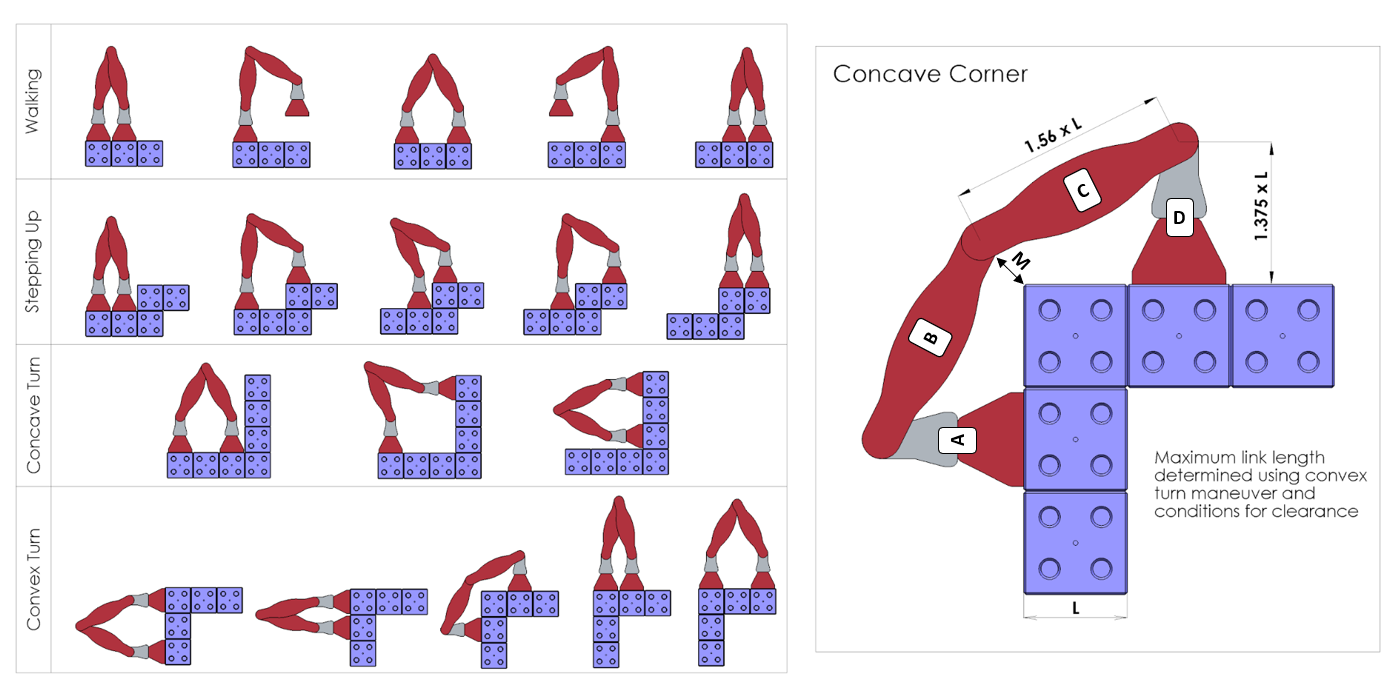}
    \caption{Navigation capabilities of the inchworm.}
    \label{fig:navigation}
\end{figure}
\textbf{Link design and analysis.} We derived a relationship between the required link lengths and the fundamental inchworm motion configurations to ensure versatile 3D navigation. Fig.\ \ref{fig:navigation} reports the fundamental motions that we considered. For this analysis, the side view of the inchworm allows us to simplify its analysis by considering a 4-link robot with links labeled A through D.
The block side length $L$ affects the lengths of links B and C. The minimum link length is constrained by the robot's ability to navigate a convex corner without collision as shown on the right of Fig.\ \ref{fig:navigation}. To determine a relationship between the inchworm link length and block side length $L$, we expressed the link lengths as vector loop equations \cite{mechanisms}. To solve these equations in the convex corner configuration, we set the length of the A and D links to \unit[1.375$L$] and the clearance vector between the middle joint and the block corner to \unit[0.25$L$]. By solving the vector loop equations, the relationship between the block side length, $L$, and the inchworm link lengths are expressed by $L_\text{A} = L_\text{D} = 1.375 L$ and  $L_\text{B} = L_\text{C} = 1.506 L$. Furthermore, if we define a worst-case scenario as a static fully extended robot that is perpendicular to the direction of gravity, as the robot becomes longer and heavier, the required joint torque increases quadratically. Based on budget considerations and a review of available customer off-the-shelf actuator options, the standard form factor servo was the strongest candidate for joint actuators. We therefore defined the block side length as $L=\unit[3]{in}$ and the current inchworm prototype is realized using servo motors. The link lengths of the inchworm were determined based on the above analysis and are reported in Fig.\ \ref{fig:inchworm_cad}. Preliminary testing of the inchworm validated that these link lengths are large enough to enable the inchworm to fully enclose the necessary electromechanical components.

\textbf{Motor selection.}
To specify a suitable motor, we calculated the dynamic torques for the worst-case scenario where each joint has a positive instantaneous velocity and acceleration. For this calculation, we considered a joint speed of \unit[30]{$^\circ$/s}, while the maximum acceleration was determined from a trapezoidal velocity profile \cite{spong_control}. Using a point mass dynamic model of the robot, the worst-case dynamic joint torque resulted in \unit[17.5]{kg-cm}. We chose a standard-form-factor JX PDI-HV5932mg \unit[30]{kg-cm} servo as it provides a safety factor of 1.5 based on the worst dynamical joint torque. We also added high-resolution absolute position magnetic encoders to the joints. 


\textbf{Gripper end effector design.}
The end effector design, depicted in Fig.\ \ref{fig:interface_cad}, consists of a flat disk with four, half-inch dowel pins integrated into the surface, as seen on the right of Fig. \ref{fig:interface_cad}. These dowel pins absorb the shear loads on the robot-to-structure interface and align the end effector with the block. We use a single, actuated 4-40 threaded rod to screw into the structure and absorb the normal load acting on the interface. The rod and pin ensure that the overall connection can sustain the loads required to be attached to the structure. Based on the dynamic model of the inchworm, the maximum torque between the inchworm and structure was calculated to be \unit[26.21]{kg-cm} during fast movements. The maximum expected load on the screw is \unit[67.6]{N}, which is well within the proof load of the gripper screw. In order to attach a block, the gripper has an actuated Allen key that is used to turn the block screwing mechanism. Lastly, each gripper has a Near-Field Communication (NFC) antenna that is used to communicate with the structure. NFC is discussed in more detail in Sec.\ \ref{sec:smartblock}.

\textbf{Electronic design.}
The inchworm uses a Teensy 3.6 micro controller and Raspberry Pi Zero for low-level control and joint trajectory planning. The AS5048B magnetic encoders were chosen for the joint sensor because of their compact size. These encoders provide a 14-bit joint angle measurement with a $\unit[0.02]{^\circ}$ resolution. Lastly, the current prototype of the robot is powered by a $\unit[9]{volt}$ power supply instead of a battery for simplicity in conducting experiments.

\begin{figure}
    \centering
    \includegraphics[width=.9\columnwidth]{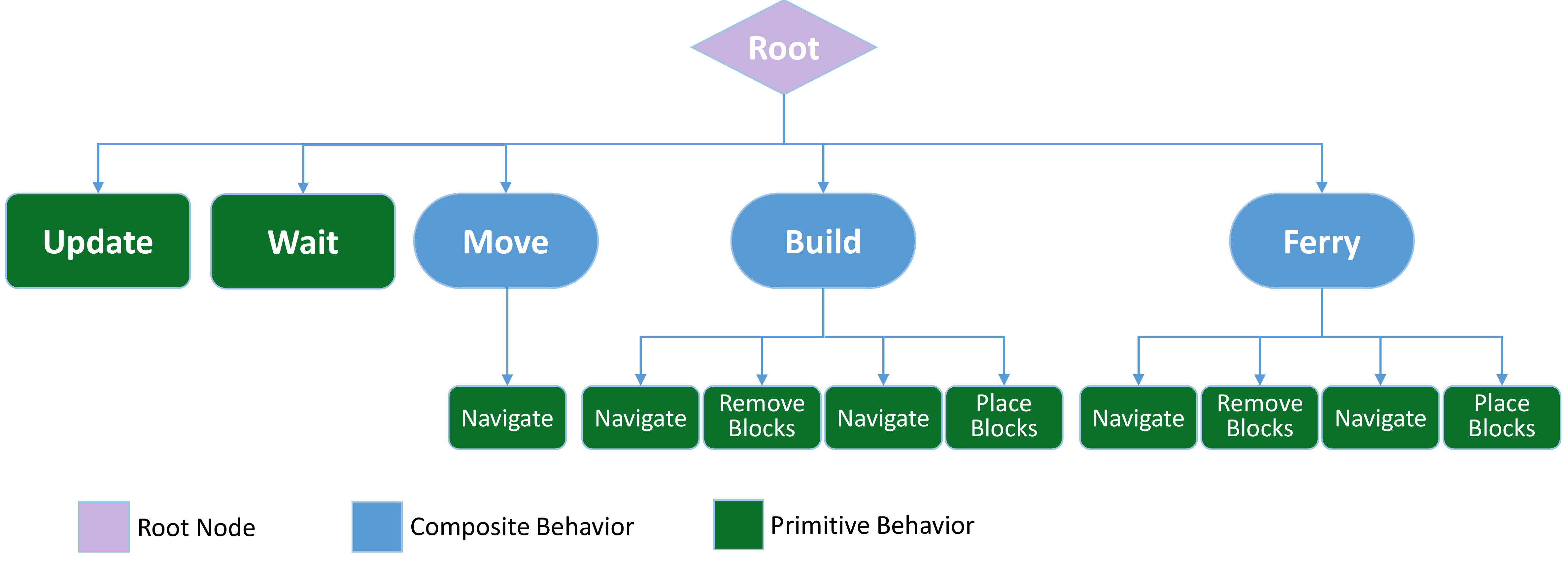}
    \caption{Inchworm robot behavior tree.}
    \label{fig:behavior_tree}
\end{figure}

\textbf{Inchworm behaviors.}
We structured the behaviors of the inchworm into a behavior tree \cite{behaviortrees}. We found that this approach fits well with the reactive nature of the inchworm behavior, while being simple and extendable. The main subtrees involve
\begin{inparaenum}[\it (i)]
\item receiving updates from the structure;
\item waiting (do nothing);
\item moving to a specified location; and
\item building sections of the structure.
\end{inparaenum}
Each of these subtrees is composed of lower-level subtrees that encode simpler behaviors such as picking up blocks, pathfinding, and placing blocks. With reference to Fig. \ref{fig:behavior_tree}, the behaviors are executed top to bottom, left to right. 

\subsection{Smart Blocks}
\label{sec:smartblock}

To address the second design goal in Sec. \ref{sec:problem_statement}, our smart blocks need to physically attach to other blocks and communicate with nearby blocks and inchworms. This way, blocks can know the current status of a structure and detect failed or detached blocks. Furthermore, an inchworm can directly communicate with a structure for localization and to determine the status of the structure.

\begin{figure}
    \centering
    \includegraphics[width=.9\columnwidth]{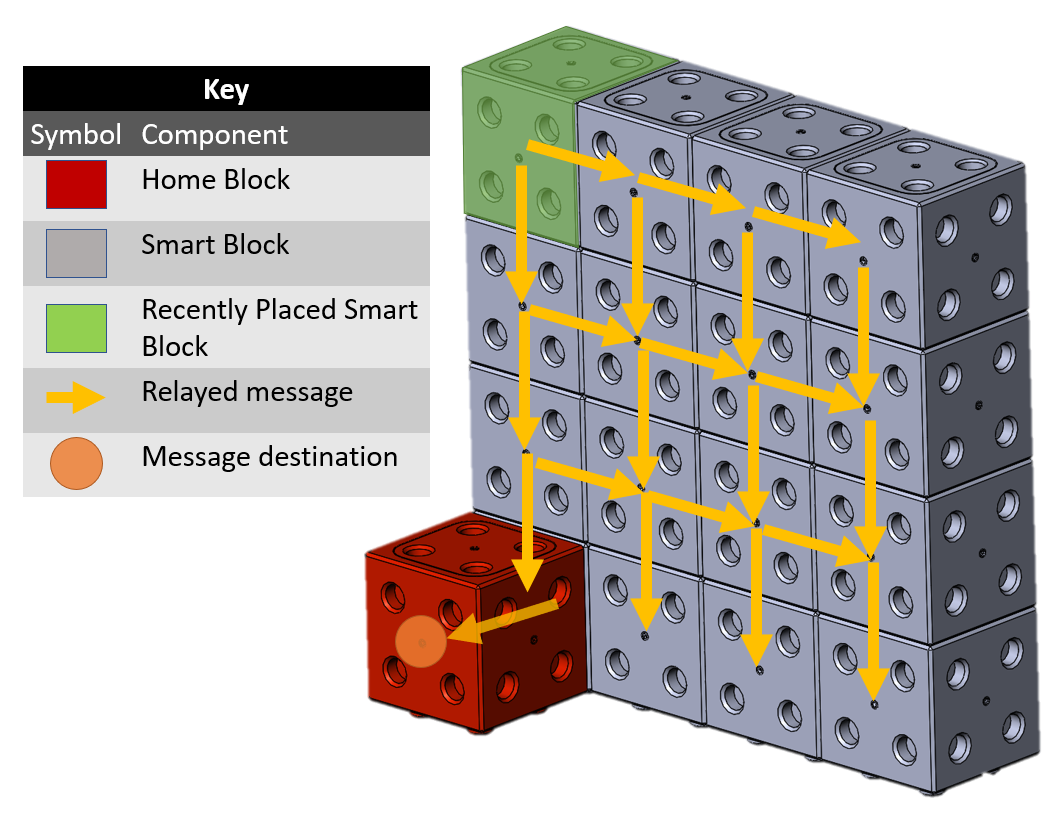}
    \caption{Structure made of smart blocks showing the flooding communication strategy.}
    \label{fig:smart_block_struct}
\end{figure}

\textbf{Communication.}
For communication to be possible across the 6 faces of a smart block, we considered multiple methods including mesh networks, Universal Asynchronous Receiver/Transmitter (UART) and NFC. We settled on an NFC-based approach with six NXP PN532 NFC controllers and antennas, one for each face, which are all controlled by an Arduino Mini microcontroller. For two separate block faces to communicate over NFC, one NFC controller must be in card emulation mode and one in card reader mode. For a block to receive messages, the NFC controller goes into card emulation mode and the microcontroller polls each block face for an incoming message. For a block to send a message, the NFC controller goes into card reader mode and the message is relayed to the receiving NFC face. The microcontroller keeps the NFC block faces in card reader mode until the message transmission has been confirmed. As shown in Fig.\ \ref{fig:smart_block_struct}, the blocks communicate through a flooding algorithm in which every block broadcasts messages on all its faces.

\textbf{Inchworm-block connections.}
Each block has five female connection faces and one male connection face. The male connection face uses a 4-40 threaded screw, like the robot gripper, to attach to a structure. The top female face has an Allen key input that the inchworm uses to operate the block screwing mechanism. This is shown in Fig. \ref{fig:interface_cad}.

\textbf{Power and LEDs.}
Each block has a 600 mAh battery which provides an average battery life of about 4 hours.
Each face on a block has programmable Neopixel LEDs. This feature is used to convey the state of the block to the user. For an example, a green block is incorporated into a structure, a flashing yellow block is waiting to be added to the structure, and a flashing red block is visually alerting the user that an adjacent block has been removed or failed.

\textbf{Home blocks.}
A variation of the smart block, the \textit{home block}, serves as the structure seed and the initial planner for the entire system. Multiple home blocks can be deployed to parallelize construction starting from several locations. The home block is equipped with a Raspberry Pi Zero.

\subsection{Structure Perception Functionality}
Three key functions leverage the symbiosis between the blocks and the inchworms:
\begin{enumerate}
    \item Structure Status: The smart blocks update the structure's current configuration as blocks are dynamically added.
    \item Robot Position: Each inchworm robot requests its own position and the position of other robots with respect to the structure, enabling localization in 3D space.
    \item Heartbeat Protocol: The structure can identify if a block is missing or has failed by regularly updating the structure’s current configuration and comparing it to the previous configuration. This is performed by checking the status of neighboring blocks.
\end{enumerate}

\section{PLANNING AND NAVIGATION}
\label{sec:planning_navigation}

To fulfill the third requirement, we devised algorithms for planning and navigation that take advantage of the hardware design presented in Sec.\ \ref{sec:robot_platforms}.

\subsection{Planning}
\label{sec:planning}

\begin{figure}
    \centering
    \includegraphics[width=.9\columnwidth]{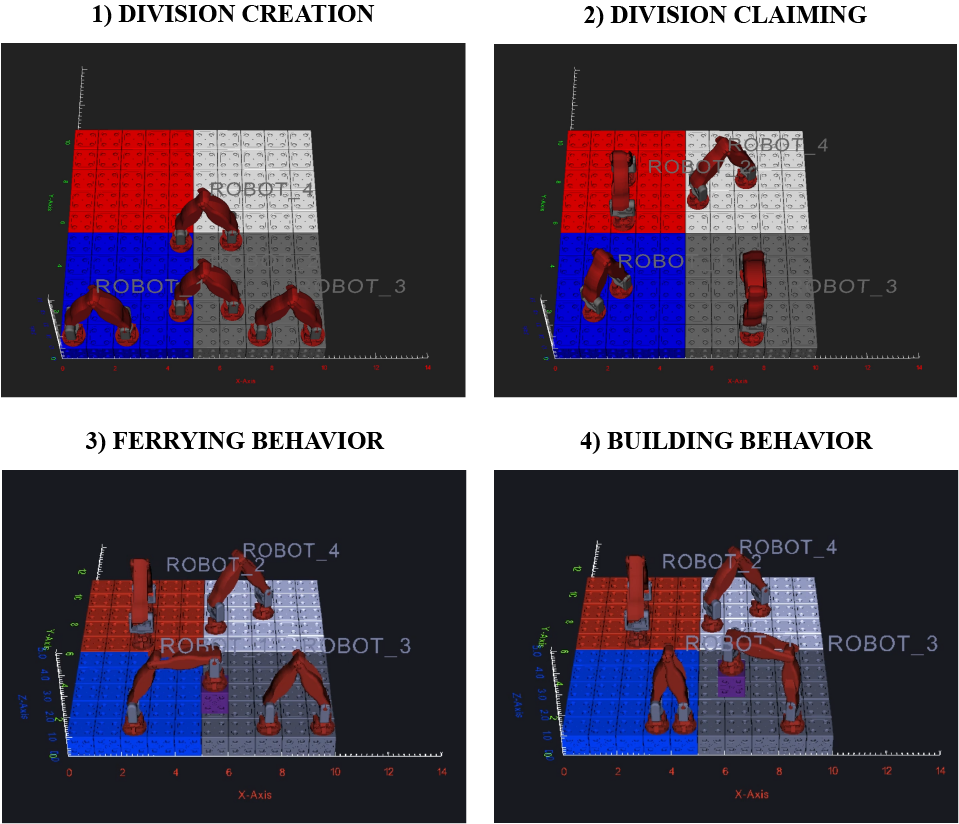}
    \caption{Examples of steps of building algorithm.}
    \label{fig:building_alg}
\end{figure}

\begin{figure}
    \centering
    \includegraphics[width=.7\columnwidth]{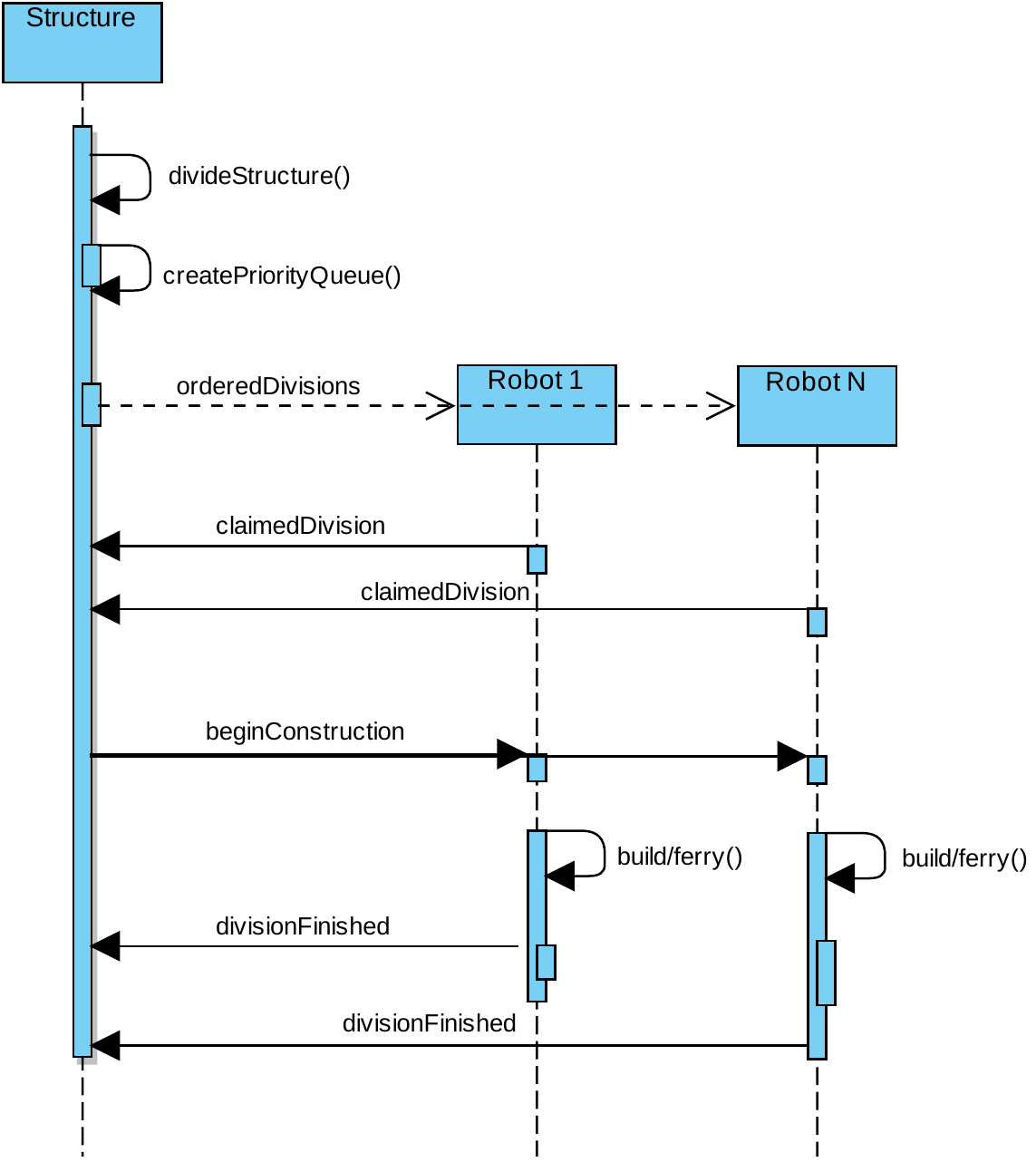}
    \caption{Sequence diagram of building algorithm for structure and N number of inchworms that is repeated for every layer within the given structure.}
    \label{fig:sequence_diagram}
\end{figure}

The planning algorithm (see Fig. \ref{fig:sequence_diagram}), is responsible for calculating the sequence of operations that must be accomplished to build the target structure. These operations include retrieving and placing blocks in the locations indicated in the blueprint of the target structure. The planning algorithm is executed by the home block (or blocks, in case multiple home blocks are deployed).

\textbf{Divisions.}
Because navigation is a time-consuming activity for the inchworms, the planning algorithm splits the structure into \textit{divisions}. Each inchworm claims a specific division. Inside of a division, the inchworm receives instructions from and communicates to the planner through the smart blocks. For example, the home block informs the inchworms that a new block is available for placement.

\textbf{Ferrying.}
When inchworms are notified that a new block is available for placement, the inchworm in the nearest division picks it up and either places it in its own division, or it brings it to the border of the division nearest to the target location. The inchworm responsible for that division picks up the block and repeats the above procedure. We call \textit{ferrying} the act of passing blocks across divisions until the target division is reached. Ferrying, akin to a bucket-brigade, is a form of task partitioning \cite{ratnieks1999task,pini2011task} that lowers the navigation cost of the inchworms. The planner provides instructions to the inchworms by changing the state of blocks.

\textbf{Division creation.}
Divisions are created by subdividing a given structure into areas where an inchworm is meant to perform work, as shown in Fig. \ref{fig:building_alg} (division creation). While the divisions extend to 3D shapes, only a layer of the structure, defined as a group of blocks 1 unit high, is considered at a time. The size and shape of divisions can vary, but ultimately they are meant to minimize the amount of inching the inchworms must perform to either ferry or build blocks. Next, a partial ordering of the divisions is established as a means to determine the order at which divisions should be built. This partial ordering is used to construct a priority queue. An assumption made as part of the system is that a location on the structure, designated as the feeding location, would be where a human operator would introduce new blocks into the system. For simplicity, the system uses a single feeding location at the home block. The partial ordering is determined as the relative distance from the feeding location and calculated using the wavefront algorithm. Divisions that are closer to the feeding location are built first. Once these divisions are built, the inchworms traverse the built divisions to ferry blocks to divisions farther away from the feeding location. 

\textbf{Division claiming.}
Once the partial ordering is established for a given layer of the structure, the inchworm robots are tasked with dispersing themselves into the different divisions, as shown in Fig. \ref{fig:building_alg} (division claiming).  The divisions are considered in the order of the priority queue. Using the communication capabilities of the smart blocks, each inchworm calculates its position in the different divisions. Each inchworm then communicates the nearest division and the distance to that division. The distance to the division is considered the amount the inchworm is bidding for a division, such that a larger distance corresponds to a higher vote. The inchworm with the highest bid (largest distance) claims the division, and the process is repeated until each inchworm has claimed a division. This process is done such that each inchworm minimizes the distance it must travel (inch) to reach a division. 

\textbf{Block placement.}
After the inchworms are distributed, the operator introduces new smart blocks into the system, which signal to the inchworms that there are new blocks to move or build. If an inchworm is in a division that has already been built, it passes the block (ferry) to another division, as shown in Fig.\ \ref{fig:building_alg} (ferrying behavior).  If the division has not been built, the inchworms query the planner for the blueprint of that division and place the blocks accordingly, as shown in Fig.\ \ref{fig:building_alg} (building behavior). When moving blocks, whether ferrying or building, the inchworms place blocks in a spiral fashion, starting from the center of the division and moving outwards. This was chosen so that after the first block is placed at the center of the division, the inchworm can climb onto the block and act as a fixed-base robot arm, pivoting around the block to place additional blocks. If the shape and size of divisions prevent the inchworm from reaching each location, the robot will inch to place a block. When ferrying, the inchworms place blocks towards the outer edges of their own division such that a robot in an adjacent division has to extend as little as possible into the initial division, thus minimizing the chances of collision or periods where an inchworm must wait for another inchworms to move. 

\textbf{Reactive replanning.}
Once the inchworms have finished building or ferrying a division, the inchworms will disperse themselves again and repeat the process of either building or ferrying blocks. This minimizes the time the inchworms spend waiting for other inchworms to finish. The procedure of dividing a layer into divisions, dispersing the inchworms, and then building or ferrying blocks through the divisions is repeated for each layer in the structure until the entire structure is built. 

\textbf{Achievable structures.}
Based off of the manner in which the planner performs construction, it is possible to characterize the types of buildings that can be built using our algorithm. Specifically, our algorithm can build any structure that does not present upside-down L-shape overhangs. This can be informally explained by noting that the building algorithm proceeds in a bottom-up (positive $z$ direction) manner, such that it will never remove a block from a layer below one that has already been built. Thus, our building algorithm is capable of building structures with overhangs, excluding those with overhangs that extend into the negative $z$ direction, characterized as an upside down L-shape. We wish to stress that this limitation concerns the building algorithm and not the hardware capabilities of the robots. Removing this algorithmic limitation is left as future work. 

\subsection{Navigation}

In order to reach any given location on the structure, the inchworm robots must be capable of producing a traversable path to the given location.

\textbf{Face*.}
We developed a simple path planning algorithm, called \textit{Face*}, that takes advantage of the versatile mobility of the robots. Face* is an extension of the A* algorithm that considers the various legal motions of the inchworms, such as the fact that it can climb vertically and upside-down, while minimizing the amount of inching performed. The algorithm works by considering the faces of each block as nodes in a network, and using the Manhattan distance to each node as the cost associated with reaching that node. Analogously to the A* algorithm, Face* selects the node with the lowest associated cost. For each node, an inverse kinematics simulation is performed to ensure no collisions with the structure are possible. The collision detection analysis is only run three times for each node to simplify complexity: once when the robot has disengaged from the current node, once when the robot has moved above the goal node, and once when the robot has engaged with the goal node. If, at any point in this process, a node is unreachable or causes a collision, the node is removed and the next lowest cost node is considered. Only when a complete, collision-free, traversable path is created does the algorithm returns a successful path. The pseudocode of Face* is shown in Alg.\ \ref{face*}.

It is important to note that a path is only generated for the end effector closest to the goal, not for both end effectors. The second end effector does not require a separate path to be generated as it can follow in the footsteps of the first end effector. Also note that if an end effector is not engaged to the structure, Face* first considers if the already disengaged end effector is capable of reaching the given position. This allows the inchworm to swing in place like a fixed-base robot arm and reduce unnecessary inching. 


\begin{algorithm}
\caption{Inchworm path planning for navigation}
\label{face*}
\begin{footnotesize}
\begin{algorithmic}[1]
\Procedure{Face*}{\textit{structure}, \textit{start}, \textit{goal}}
    \State \textit{frontier} $\gets$ PriorityQueue((\textit{start}, 0))
    \State \textit{cameFrom} $\gets$ $\{(None, 0)\}$

    \While{\textbf{not} \textit{frontier}.empty()}
         \State \textit{current} $\gets$ \textit{frontier}.get()
         
         \If{\textit{current} == \textit{goal}}
            \State \textbf{break}
         \EndIf
         
         \For{\textit{next} \textbf{in} \textit{structure}.neighbors(\textit{current})}
            \State \textit{newCost} $\gets$ \textit{cameFrom}[\textit{current}[1] + \textit{structure}.cost(\textit{current}, \textit{next})
            \If{\textit{next} \textbf{not in} \textit{cameFrom} \textbf{or} \textit{cameFrom}[1] \textless  \textit{\textit{current}}[\textit{next}]}
            \If{\textit{structure}.isReachableAndCollisionFree(\textit{next})}
                \State \textit{cameFrom}[\textit{next}][1] $\gets$ (\textit{current}, \textit{newCost})
                
                \State \textit{priority} $\gets$ \textit{newCost} + heuristic(\textit{goal}, \textit{next})
                
                \State \textit{frontier}.put(\textit{next}, priority)

         \EndIf
                
            \EndIf
         
         \EndFor

    \EndWhile
    \State \textbf{return} \textit{path}
\EndProcedure
\end{algorithmic}
\end{footnotesize}
\end{algorithm}

\section{EXPERIMENTAL EVALUATION}
\label{sec:experimental_evaluation}

To gauge the performance of our approach, we ran both simulated and real-world experiments.

\subsection{Simulated Experiments}

We developed a custom simulation environment to visualize and collect statistics of the system. The simulation environment was built using the Visualization Tool Kit (VTK).\footnote{\url{https://vtk.org/}} The experiments were run on a 2015 MacBook Pro.

\textbf{Metrics.} The performance metrics we considered for the simulated experiments are
\begin{inparaenum}[\it (i)]
\item whether the construction process succeeded; and
\item the speed at which the structure was constructed, measured in simulated time steps.
\end{inparaenum}

\begin{figure}[t]
    \centering
    \includegraphics[width=.8\columnwidth]{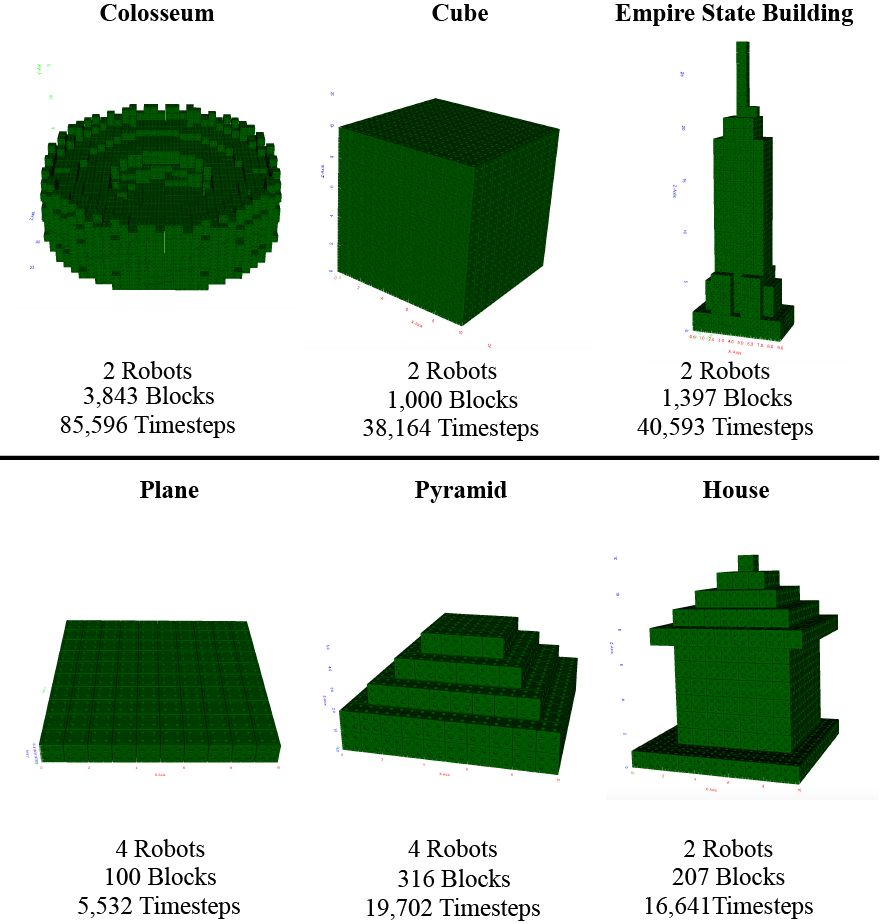}
    \caption{Examples of structures achievable by our system.}
    \label{fig:simulated_buildings}
\end{figure}

\textbf{Setup.}
We tested our approach with diverse structures including pyramids, temples, churches, castles, the Empire State Building, and the Star Trek Reliant ship, among others (see Fig.\ \ref{fig:simulated_buildings}). We compared the construction speed across different numbers of inchworms to investigate the effect of adding additional robots on construction time. In the simulation experiments, inchworms were randomly placed on a surface composed of smart blocks. The inchworms were then tasked with distributing themselves according to the distribution algorithm explained in Sec.\ \ref{sec:planning}. After the inchworms had distributed themselves, a virtual operator sets smart blocks down in a corner of the layer, designated as the feeding location. Prompted by the introduction of a new block into the system, the inchworms reacted by moving the blocks throughout the system as explained in Sec.\ \ref{sec:planning_navigation}. For all simulations, the inchworms were able to place all blocks as required.

\begin{figure}
    \centering
    \begin{subfigure}{.9\columnwidth}
        \includegraphics[width=.9\columnwidth]{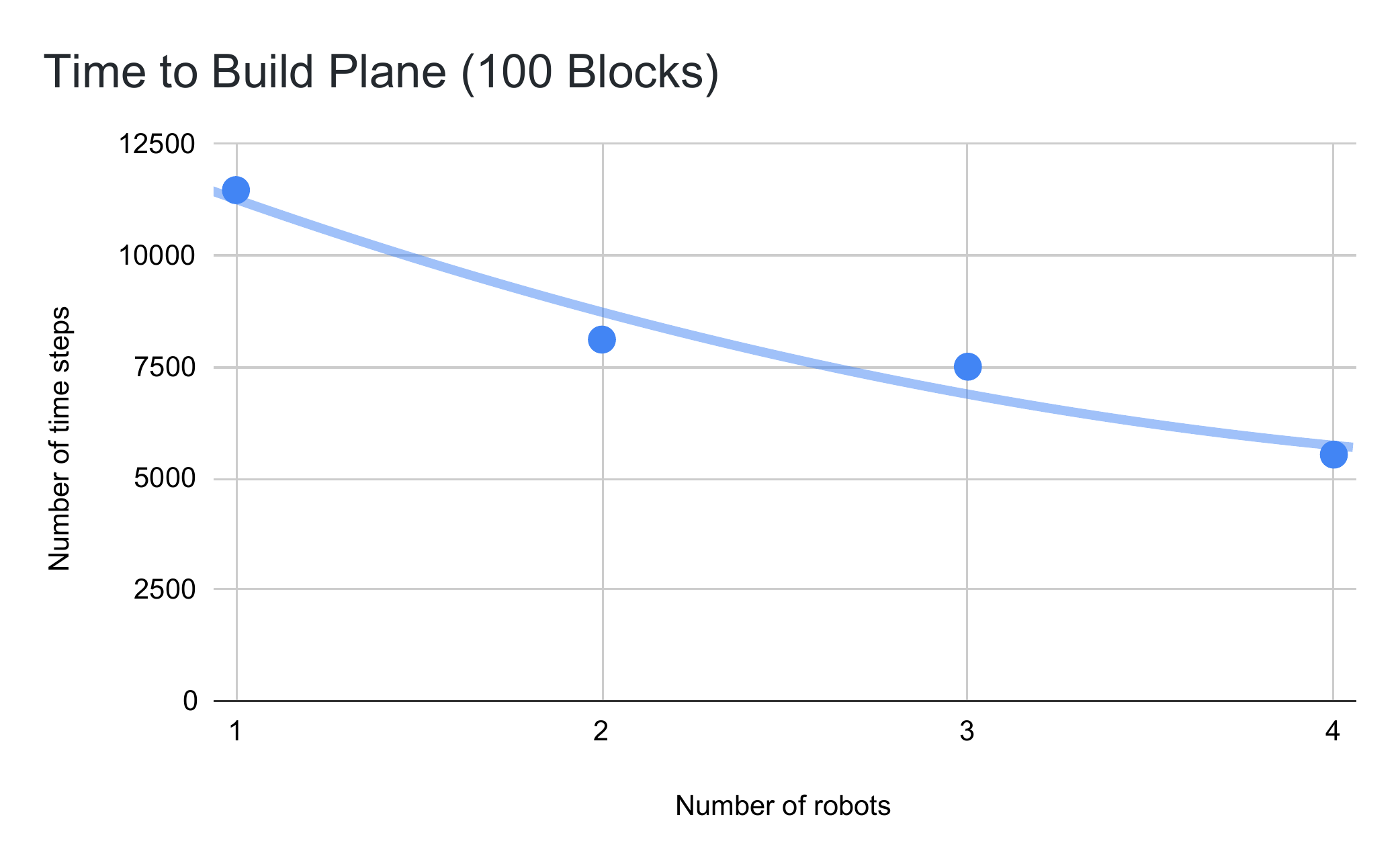}
        \label{fig:sub1}
    \end{subfigure}
    \begin{subfigure}{.9\columnwidth}
        \includegraphics[width=.9\columnwidth]{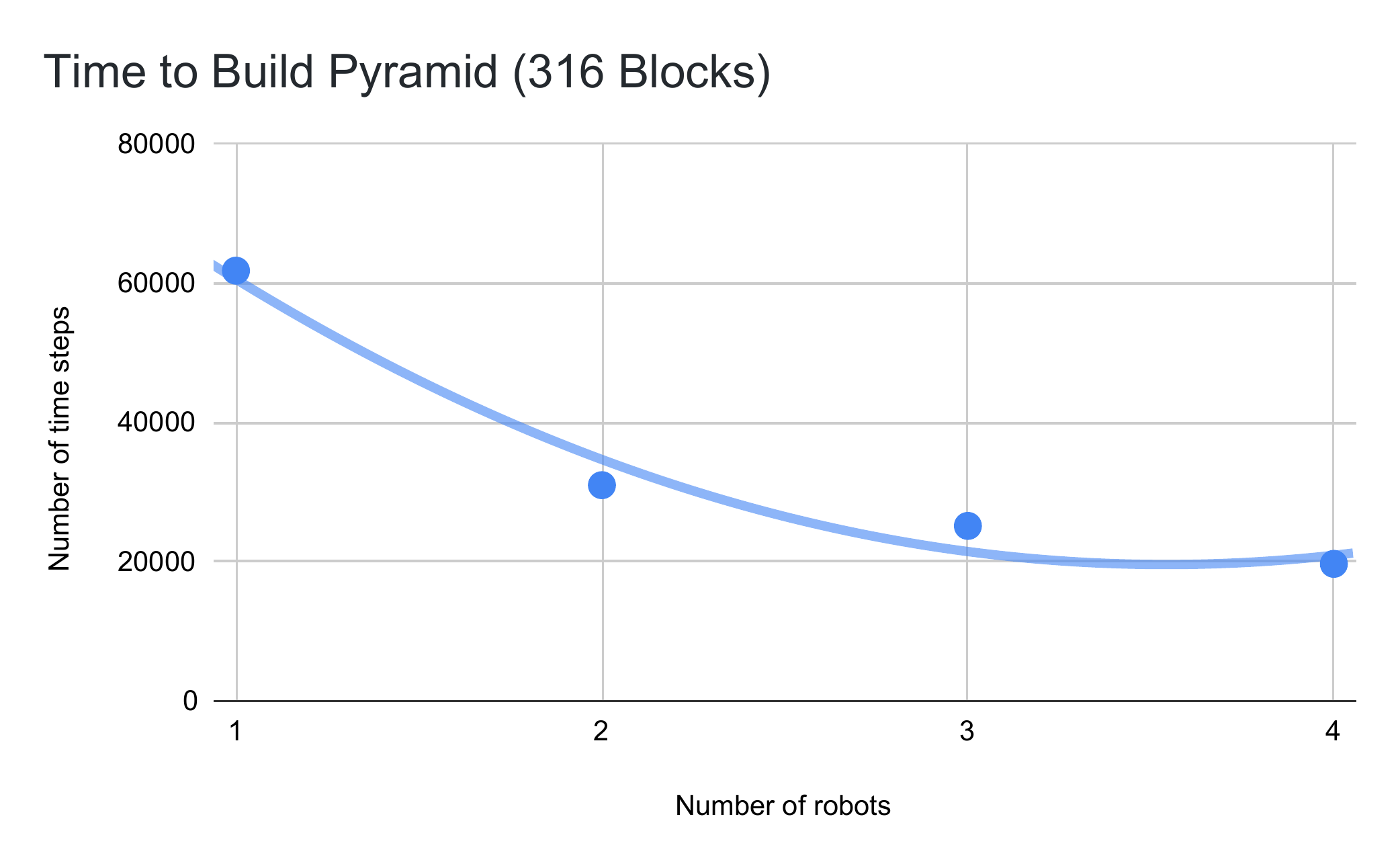}
        \label{fig:sub2}
    \end{subfigure}
    \caption{Time required to build structures for different numbers of inchworms.}
    \label{fig:results_plane}
\end{figure}

\textbf{Scalability study.}
The results of two experiments are reported in Fig.\ \ref{fig:results_plane}. The experiments involve 1 to 4 robots tasked with building a $10 \times 10$ plane and a pyramid consisting of 316 blocks. As shown in Fig.\ \ref{fig:results_plane}, the time decreases with additional robots, especially when comparing a single robot and four robots. The results appear to follow a sublinear trend, which we approximated with a second-order polynomial curve.

\begin{figure}
    \centering
    \includegraphics[width=.7\columnwidth]{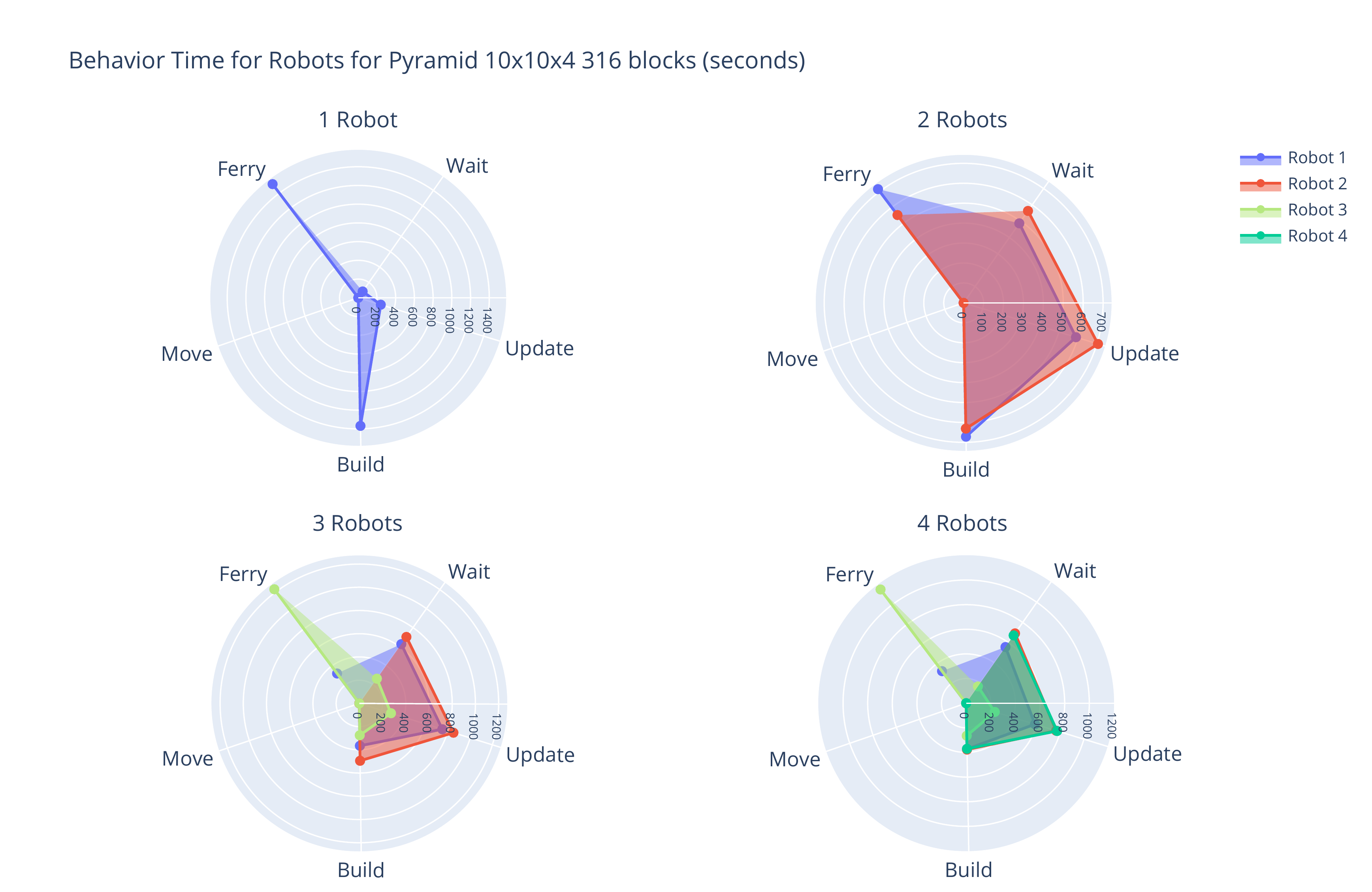}
    \caption{Time spent performing specific behaviors for different numbers of robots.}
    \label{fig:behavior_time}
\end{figure}

\textbf{Task specialization.}
To analyze how the collaboration between inchworms affects performance, we recorded the total amount of time spent in each behavior by the inchworms. The main behaviors tracked were receiving updates from other robots and the structure (\textit{Update}), moving (\textit{Move}), waiting (\textit{Wait}), building (\textit{Build}), and ferrying (\textit{Ferry}). The results shown in Fig.\ \ref{fig:behavior_time} refer to the same structures considered in the scalability study. For the single inchworm experiment, the inchworm spends almost all its time either ferrying or building. Almost no time is spent waiting, and a very small portion of time is spent receiving updates. This is expected, as there are no other inchworms with which it must coordinate.
For the experiments with multiple robots, we observed that the time spent across the different behaviors is very heterogeneous. Although capable of exhibiting all the behaviors, the inchworms specialized themselves into different groups, in which some of the inchworms focus more on ferrying while others focus more on building. In particular, we observed that one inchworm did most of the ferrying, while the other inchworms mainly focused on the building. This task specialization is an interesting emergent property of our approach whose appearance depends on the type of structure to build. We also observed that the waiting time and the time spent receiving updates of all inchworms is much higher compared to the single inchworm experiment. This is expected, as by increasing the number of inchworms, maintaining coordination demands communication between the inchworms and the smart blocks. Samples of these tests are reported in the video at \url{https://youtu.be/zLDa8vK0HpM}.


\subsection{Real-world Experiments}
The goal of the real-world experiments was to validate the hardware, as well as to investigate whether some of the results found in the simulation would be consistent in the real world. We constructed one inchworm and five smart blocks.

\textbf{Unit testing.}
We performed a successful campaign of unit tests to validate the fundamental motions of the inchworm. We also tested the smart blocks to detect when new blocks were added or removed, or when a block went offline, which could be attributed to a fault, dead battery, or other physical error. The tests were all successful. Samples of these tests are reported in the video at \url{https://youtu.be/zLDa8vK0HpM}.

\textbf{Inching precision test.}
We validated the precision of the inchworm to estimate how often an inchworm may fail. The inchworm was made to take an inching step and retract 10 times. This experiment was also replicated while an inchworm was taking a step with a block. It was found that the inchworm was able to successfully step and grip 100\% of the time and step while carrying a block 80\% of the time. The results from the second test is attributed to the disturbance the block applies on the inchworm which can cause occasional inching errors.

\textbf{Inching and gripping test.}
We also tested the capability of our inchworm to take three steps on a plane while carrying a block. We estimated the inching speed to be 1 step every \unit[40]{s}. The majority of this time came from the grippers, with about \unit[20]{s} on average for engaging with the structure and about \unit[10]{s} for disengaging with the structure.

\begin{figure}
    \centering
    \includegraphics[width=.9\columnwidth]{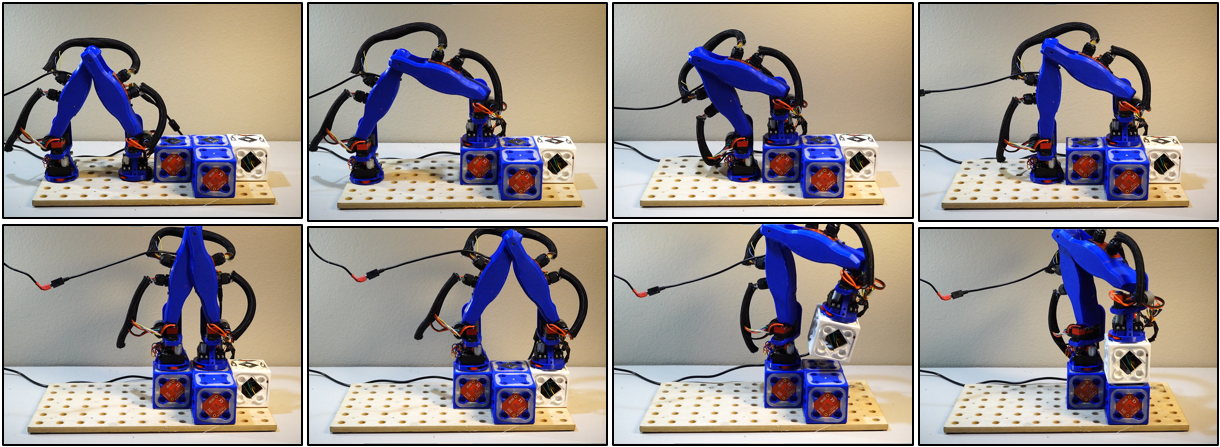}
    \caption{Robot inching and placing a block.}
    \label{fig:system_inch_test}
\end{figure}

\textbf{Inchworm system test.}
A complete system-level test was also performed to demonstrate that the inchworm is able to inch, traverse a structure, manipulate a block, and place it on the structure. A series of snapshots from the demonstration are shown in Fig. \ref{fig:system_inch_test}, and the footage is available in the video at \url{https://youtu.be/zLDa8vK0HpM}.


\begin{figure}
    \centering
    \includegraphics[width=.9\columnwidth]{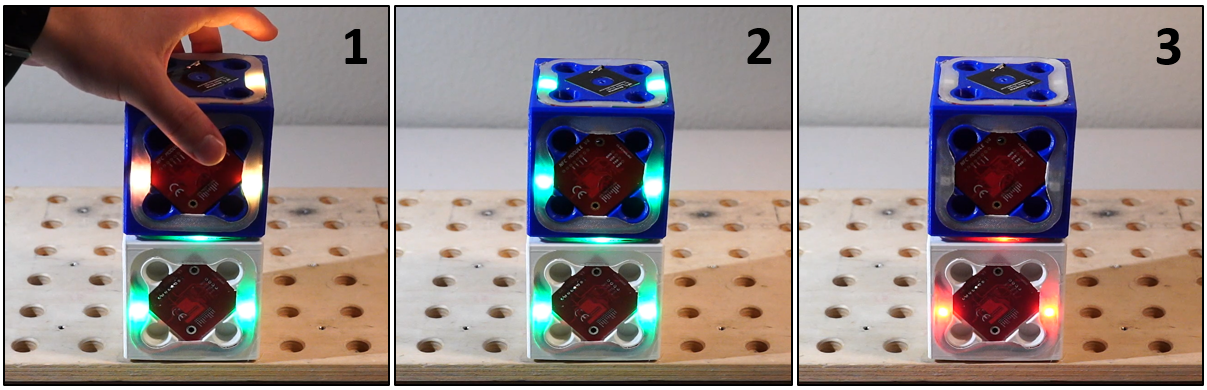}
    \caption{
    1) A block is added to the structure. 2) The top block turns green to indicate ``all OK.'' 3) The bottom block changes color to red after it detects an error in the top block, which has gone offline.}
    \label{fig:block_test}
\end{figure}

\textbf{Smart block communication.}
We performed a series of tests to validate that a new block is detected when it is added into the structure, and that the structure can detect when a block is removed from the structure. The images in Fig.\ \ref{fig:block_test} show that the blocks change the color of their LEDs to convey the change in the structure. Experimentally, the average time for two smart blocks to exchange a 62-byte package was \unit[500]{ms}, which corresponds to an average bit rate of \unit[992]{bit/s}. We also estimated the impact of a large structure on communication latency. The number of hops between any block and the home block is given by the Manhattan distance between them. A message originating from block at $(x,y,z)$ of a 3D structure must traverse $x+y+z$ blocks to reach a home block at $(0,0,0)$. To put this in perspective, a message sent by a block placed in the corner of a $10 \times 10 \times 10$ cube would take \unit[15]{s} to reach the opposite diagonal corner. This has important implications when building large structures, motivating the need for multiple home blocks to decrease this latency. We will study these aspects in future work.

\section{CONCLUSIONS AND FUTURE WORK}
\label{sec:conclusions}

This paper proposes \smac, a robotic platform for collective construction composed of two types of robot:
\begin{inparaenum}[\it (i)]
\item a collection of smart blocks, forming an intelligent network capable of dynamic planning and monitoring; and
\item an inchworm-shaped, builder robot, able to navigate a 3D structure and deposit blocks under the guidance of the smart blocks.
\end{inparaenum}
We presented the hardware design of our platform, along with a set of simulated and real-world experiments aimed at showing its capabilities.

Our work is a step in the study of ``active'' stigmergy in multi-robot construction, a promising concept in which the structure being built is an active component of the process. This idea makes it possible to simplify the design of the builder robots. In addition, the smart blocks could be used as a form of \textit{smart scaffolding}. This would make it possible to reuse the same robotic platforms in multiple construction projects, thus paving the way for affordable, cost-effective construction in space applications and remote terrestrial environment.

Future work will concentrate on improving the capabilities of our robotic platforms and devising/studying novel planning algorithms to exploit the concept of active stigmergy in collective construction.

\section*{ACKNOWLEDGMENT}

We thank Prof.\ Raghvendra Cowlagi for guidance and fruitful conversations and Cameron Collins for his assistance in the manufacturing of the hardware of the platform.

\bibliography{bibliography}
\bibliographystyle{IEEEtran}

\end{document}